# A Joint Probabilistic Classification Model of Relevant and Irrelevant Sentences in Mathematical Word Problems

SULEYMAN CETINTAS AND LUO SI

Department of Computer Sciences,
Purdue University, West Lafayette, IN, 47907, USA
{scetinta,lsi}@cs.purdue.edu
and
YAN PING XIN, DAKE ZHANG AND JOO YOUNG PARK

Department of Educational Studies,
Purdue University, West Lafayette, IN, 47907, USA
{yxin,zhang60,park181}@purdue.edu
and
RON TZUR

Department of Mathematics Education
University of Colorado Denver, Denver, CO, 80217, USA
ron.tzur@ucdenver.edu

______________________________________________________________________

Estimating the difficulty level of math word problems is an important task for many educational applications. Identification of relevant and irrelevant sentences in math word problems is an important step for calculating the difficulty levels of such problems. This paper addresses a novel application of text categorization to identify two types of sentences in *mathematical word problems*, namely *relevant* and *irrelevant* sentences. A novel joint probabilistic classification model is proposed to estimate the joint probability of classification decisions for all sentences of a math word problem by utilizing the correlation among all sentences along with the correlation between the question sentence and other sentences, and sentence text. The proposed model is compared with i) a SVM classifier which makes independent classification decisions for individual sentences by only using the sentence text and ii) a novel SVM classifier that considers the correlation between the question sentence and other sentences along with the sentence text. An extensive set of experiments demonstrates the effectiveness of the joint probabilistic classification model for identifying relevant and irrelevant sentences as well as the novel SVM classifier that utilizes the correlation between the question sentence and other sentences. Furthermore, empirical results and analysis show that i) it is highly beneficial not to remove stopwords and ii) utilizing part of speech tagging does not make a significant improvement although it has been shown to be effective for the related task of math word problem type classification.

______________________________________________________________________

## 1. INTRODUCTION

Math performance on high-stakes tests has become increasingly important in recent years and there has been some improvement in academic achievement following the passage of the No Child Left Behind Act in 2002 [US. Dept. of Education, 2006]. Yet, the overall performance of American students in math has been of particular concern [Gollup et al. 2002]. Increasing trends in computers' utilization for math teaching have led to the development of various intelligent tutoring systems (ITS) in the math domain [Beal, 2007; Koedinger et al., 1997; Shiah et al., 1995].

Intelligent Tutoring Systems (ITS) provide their effective, individualized instruction by adjusting the characteristics (e.g., type, difficulty level, context) of their educational materials dynamically for each student. For instance, if a student is found to benefit from a specific difficulty level, then such a problem should be available. Similarly, if a student seems bored of a particular problem, the problem type or context can be adjusted according to the student's interests. If a system runs out of educational materials of a particular kind, the benefits of ITS become limited. Therefore, providing sufficient content is very essential for ITS [Arroyo et al., 2001; Arroyo and Woolf, 2003; Birch and Beal, 2008; Hirashima et al., 2007; Ritter et al., 1998]. However, providing sufficient educational content is a time consuming, labor intensive process that requires expert knowledge of the domain, and has long been seen as a major bottleneck for the development of tutoring systems [Arroyo et al., 2001; Arroyo and Woolf, 2003; Birch and Beal, 2008; Hirashima et al., 2007; Ritter et al., 1998]. There have been some attempts to reduce the development costs by bringing teachers and students into the process of content creation (specifically, math word problem creation) [Arroyo et al., 2001; Arroyo and Woolf, 2003; Birch and Beal, 2008; Hirashima et al., 2007]. However, teachers are extremely busy people, and authoring process makes their work even harder instead of easing their load [Arroyo and Woolf, 2003]. Having students create math word problems (i.e., content) from available information is known as *problem posing,* and have been noted to reduce the development time significantly (as much as creating 60+ math word problems from 2 days to 2 hours [Birch and Beal, 2008]). Yet, created content needs reviewing in terms of language, accuracy, quality, difficulty level, etc. [Arroyo and Woolf, 2003; Birch and Beal, 2008].

Problem posing not only helps intelligent tutoring system developers accelerate content authoring, but is well-recognized as an important way to learn mathematics [Brown and Walter, 1990; Hirashima et al., 2007]. However, it has explicitly been noted that analysis of student created problems puts a very high burden on teachers, meaning that problem posing could become less popular as a learning method in reality despite its importance [Hirashima et al., 2007]. According to these authors, to evaluate student posed problems, the type of the created problems should be detected, important information (i.e., relevant information for the solution of the problem) in the sentences should be found, and relations between those should be identified. They proposed a problem posing environment with sentence integration (i.e., individual sentences of the problems were provided to students to arrange) that could analyze the student created problems automatically with its sentence integration framework. However it does not allow students to create problems directly, therefore does not help to analyze the fully-student-generated problems automatically. Cetintas et al.*[2009],* recently attempted to solve the first task for the analysis of user-generated problems, namely the identification of math word problem types. However, the identification of relevant and irrelevant information, as the second step, still remains to be solved. An early work that is relevant to this task was done in [Bobrow, 1964]. He developed the STUDENT system that accepts high school algebra word problems from its users, analyzes the user-asked problems automatically, and answers the questions based on the information contained in the inputted problems. The STUDENT system, which has been manually designed by exploiting the structure of the high school algebra problems, tries to interpret the sentences of problems with format matching. Although the STUDENT system has been shown to be effective for most of the algebra word problems in the first year high school text books, it is noted that it would not handle problems with excessive verbiage or implied information [Bobrow, 1964]. Therefore the need to identify relevant and irrelevant information still remains to be resolved. Another related work that focuses on relevant and irrelevant information was recently done by Cetintas et al., who proposed to automatically identify students' relevant and irrelevant questions asked in a micro-blogging supported classroom by utilizing the correlation between questions and available lecture materials [Cetintas et al., 2010; Cetintas et al.*,* To Appear]. Yet, their work focused on classifying the questions asked in a lecture, which is a different task from identifying the relevant and irrelevant sentences in a math word problem, and they did not use a joint classification technique that considers the correlation among all questions.

In mathematics education, *word problem* or *story problem* is the term that is often used to refer to any mathematical exercise on which significant background information is presented as text rather than in mathematical notation [Verschaffel, 2000]. The focus of this paper is a novel application of text categorization to identify two sentence types in an arithmetic word problem, namely *relevant* and *irrelevant* sentences (examples given in Table I). A *relevant sentence* in a math word problem is a sentence that contains useful information for the solution of the problem, whereas an *irrelevant sentence* is non-informative. Students with math learning disabilities have difficulties in solving all types of word problems but especially the word problems with irrelevant information [Marzocchi et al.*,* 2002; Mastropieri and Scruggs, 2006; Passolunghi and Siegel, 2001].

Table I. A Math Word Problem Example with Relevant and Irrelevant Sentences. The irrelevant sentences that are shown in italics are not informative for the solution of the problem. Note that the stopwords are in bold.

> *The Island Tours Theater opens **except for** Sunday. **The** theater opens only in the morning.* **There are** 210 people **on a** tour. **The** tourists **are** divided **into** equal groups. **If each** group **has** 7 people, **how many** groups **will there be**?

Therefore, the level of noisy data (i.e., the number of irrelevant sentences and non-informative numbers in those sentences) is one of the factors in determining the difficulty level of the math word problem, along with other factors such as the readability level. This paper studies the classification problem of relevant and irrelevant sentences in mathematical word problems and aims to be a component of math tutoring systems i) that need to automatically construct libraries of word problems, and ii) that aim to make problem posing an easy-to-use technique by automating the analysis of student-generated problems. Unlike traditional text categorization that only considers the text in the documents to be classified, relevant and irrelevant sentences in math word problems are short and correlated with each other as they together form the math word problem. The correlation between those sentences can be utilized to improve the effectiveness of the categorization.

This paper proposes a novel joint probabilistic classification model to estimate the joint probability of the classification decisions of all sentences in a math word problem. The proposed approach that utilizes the correlation between all sentences in a math word problem along with the correlation between the question sentence and other sentences and the sentence text, is compared to i) a traditional text categorization approach that only considers sentence text and ii) a novel text categorization approach that utilizes the correlations between the question sentence and other sentence in a math problem along with the sentence text. We show that i) the approach of utilizing the correlation between the question sentence and other sentences along with the sentence text significantly outperforms the traditional text categorization approach of only using the text in sentences, ii) the approach that additionally considers the correlation between all sentences outperforms both of the aforementioned approaches. Furthermore, the paper explores the effect of using stopword elimination and part of speech tagging in all the models. Experiment results show that including stopwords i) into the feature space of SVM classifiers and ii) into the feature space to be used for estimating the correlation between sentences, significantly improve the effectiveness of the classifiers. Moreover, utilization of part of speech tagging has not been found to result in consistent and significant improvements in this work, although it has been shown to be beneficial in prior work for the related task of categorizing math word problems with respect to their types [Cetintas et al., 2009].

## 2. METHODS AND MODELING APPROACHES

This section describes several modeling approaches for the categorization of relevant and irrelevant sentences in mathematical word problems.

### 2.1 Independent Classification Model (SVM_TermsOnly)

A standard text categorization (TC) model, which will be referred as the independent classification model, makes the classification decisions of documents independently by only considering the individual features of the documents (i.e. bag of words representation of the sentences in math word problems). In this work a SVM classifier is chosen as the independent classification model[1] since it is one of the most accurate and widely used text categorization techniques. Particularly the simplest linear version of SVM is chosen since it is fast to learn and fast to classify new instances [Joachims, 1998; Yang and Liu, 1999]. The SVM model in this work can be formulated as a solution to an optimization problem as follows:

$$\min_{w,b} \frac{1}{2}\|w\|^2 + C \sum_{i=1}^{N} \xi_i \quad (1)$$

$$subject\ to\ y_i(\vec{w} \cdot \vec{d_i} + b) - 1 + \xi_i \geq 0 \quad \forall\ \xi_i \geq 0$$

where $\vec{d_i}$ is the $i^{th}$ document represented as a bag of words in the TC task; $y_i \epsilon \{-1, +1\}$ is the binary classification of $\vec{d_i}$, $\vec{w}$ is the parameters of the SVM model and b is the bias parameter. $C$ has the control over the tradeoff between classification accuracy and margin, which is tuned empirically. The categorization threshold of each SVM classifier is learned by 2-fold cross validation in the training phase (i.e., sentences from half of the training problems are used for the first-fold while sentences from the problems in the other half are used for the second-fold).

This independent model is a standard TC method, as mentioned, and will be referred to as SVM_TermsOnly.

---

[1] The SVM[light] toolkit [Joachims, 1999] is utilized.

## 2.2 Improved Classification Model (SVM_TermsQSSim)

The independent classification model neglects the valuable information that is hidden in the correlations among sentences, especially the correlation between a question sentence and other sentences in a math word problem. As the question sentence depends on the information given in the relevant sentences, it is important to investigate the correlation between individual sentences and the question sentence. This sets the grounds for the motivation of this improved classification model.

We propose an improved version of the SVM_TermsOnly classifier by incorporating the correlations between all sentences in a problem with respect to the question sentence. The correlation between the question sentence and other sentences are calculated via the common cosine measure [Baeza-Yates and Ribeiro-Neto, 1999] as follows:

$$Sim(S_i, S_j) = cos(\vec{S_i}, \vec{S_j}) = \frac{\vec{S_i} * \vec{S_j}}{\|\vec{S_i}\|\|\vec{S_j}\|} \quad (2)$$

where $S_i$ is the i$^{th}$ sentence and $S_j$ is the question sentence in a problem, $\vec{S_i}$ and $\vec{S_j}$ are sentences represented as the bag of words vectors, and "$*$" denotes the dot product of these vectors. We use the common tf-idf weighting scheme along with the cosine measure to calculate the sentence to question sentence similarities [Baeza-Yates & Ribeiro-Neto, 1999]. Tf-idf weighting scheme uses term frequency (i.e., frequencies of terms in a sentence) and inverse document frequency (that favors discriminative terms that only reside in a small number of sentences). Adding these similarity scores into the baseline SVM classifier as a new dimension of the feature space enables us to use the correlation between the question sentence and other sentences in this improved model, which will be referred to as SVM_TermsQSSim. The categorization threshold of this classifier is also learned by 2-fold cross validation in the training phase.

## 2.3 Joint Probabilistic Classification Model (JointProbClass_Model)

Utilizing the correlation between the question sentence and other sentences in a problem is an effective way of improving the classification performance over a standard TC method with an independent classification model. However, there is still more room to improve. In a math word problem, there is often some correlation between a relevant sentence with another relevant sentence as well as an irrelevant sentence with another irrelevant sentence. This is intuitive as all relevant sentences are giving information for the final question that the problem is asking while most irrelevant sentences tend to talk about the same irrelevant concept.

The proposed joint probabilistic classification model considers the classification score from the SVM_TermsQSSim classifier on every sentence as well as the correlations among all sentences to estimate the joint probability of the classification decisions of all sentences via a probabilistic graphical model.

Applications of graphical models have been used to solve problems in many different domains. As undirected graphs can be used to represent correlations between variables [Cowell et al., 1999], an undirected probabilistic graphical model is used in this work for the joint classification model. A Boltzmann machine [Hinton and Sejnowski, 1986] is a special type of undirected graphical models that has been applied to tasks such as question answering [Ko et al., 2007]. The proposed joint probabilistic classification model is a version of a Boltzmann machine adapted for text classification and can be seen in detail in Figure 1.

$$P(S_1, \ldots, S_N) = \frac{1}{Z} \exp \left\{ \begin{array}{l} \alpha_0 \sum_{j=1}^{N}[SVM(S_j)\delta(S_j == 0)] + \alpha_1 \sum_{j=1}^{N}[SVM(S_j)\delta(S_j == 1)] + \\ \beta_{11} \sum_{k} \sum_{k \neq j}[Sim(S_k, S_j)\delta(S_k == 1)\delta(S_j == 1)] + \\ \beta_{00} \sum_{k} \sum_{k \neq j}[Sim(S_k, S_j)\delta(S_k == 0)\delta(S_j == 0)] \end{array} \right\}$$

Fig. 1. Joint Probabilistic Classification Model

$SVM(S_i)$ is the classification score from the SVM_TermsQSSim classifier for the i[th] sentence in a problem and is used to produce a sentence relevance score for an individual sentence; $Sim(S_i, S_j)$ is the similarity score between a sentence $S_i$ and another sentence $S_j$ in a problem and represents the correlation among sentences; $\delta(S_i == 1)$ is an indicator function and is 1 if the sentence $S_i$ is relevant, 0 otherwise; $\delta(S_i == 0)$ is again an indicator function and is 1 if the sentence $S_i$ is irrelevant, 0 otherwise; $\alpha_0, \alpha_1, \beta_{11}, \beta_{00}$ are model parameters and are estimated from the training data by maximizing the log likelihood using the Quasi-Newton algorithm [Minka, 2003]. As can be seen in Figure 1, the joint probabilistic classification model not only utilizes the correlation between the question sentence and other sentences along with the sentence text (i.e., by including $SVM(S_i)$ scores), but also considers the correlation among all sentences (i.e., by incorporating $Sim(S_i, S_j)$ scores). The joint model helps to make a more accurate decision for the sentences whose SVM_TermsQSSim scores are not accurate (i.e., leading to wrong categorization) but have high similarity with sentences of the same class, as high similarity scores will compensate for the total sum in the exponential in Figure 1.

For a test problem of N sentences, all possible configurations that consider sentences as relevant or irrelevant are calculated (i.e. each sentence can be relevant or irrelevant, making a total of $2^N$ configurations to consider for N sentences) with the joint probabilistic model using the estimated parameters. The configuration with the highest probability gives the joint probabilistic classification decision for all sentences of that problem. Note that calculation of all configurations is a burden if the number of nodes in the graph (i.e., number of sentences to jointly classify) is large. In a math word problem, there are less than 10 sentences which is a key point that makes it possible to model a joint probabilistic classifier, unlike traditional classification tasks where there are too many documents to consider and joint classification is intractable.

The joint probabilistic classification model will be referred to as JointProbClass_Model.

## 2.4 Avoiding Stopwords' Elimination

As a common text preprocessing technique, stopword removal, suggests that many of the most frequent terms in English such as *why, where, he, she, there, is, etc.* are not content words as they appear almost in every document, do not carry important information about the context of the documents and should be removed [Frakes and Baeza-Yates, 1992; Scott and Matwin, 1999; Sebastiani, 2002]. However, prior research has also shown that stopwords can be useful for text categorization of mathematical word problems (i.e. with respect to their types) [Cetintas et al., 2009]. In this paper, the effect of avoiding stopwords' removal over classifier performance is explored in two ways: i) when they are not removed for the bag of words representation of the input space for the SVM classifiers, and ii) when the correlations among sentences are calculated with the cosine similarity measure. We used the Lemur information retrieval toolkit[2] for stopwords' removal and stemming. Stemming is applied to the data for all the models in this work. In particular the INQUERY stopwords list and Porter stemmer were utilized respectively [Porter, 1980].

## 2.5 Utilizing Part of Speech (POS) Tagging

Part of speech (POS) tagging is the process of identifying the corresponding linguistic category (e.g. noun, verb, adjective, adverb, etc.) of words in a text. In recent prior research, part of speech tagging has been shown to be useful for the classification of mathematical word problem with respect to their types [Cetintas et al. 2009]. The main intuition for utilizing the part of speech tagging is the fact that different parts of speech can be discriminative for different classes in classification tasks. In this work, the effect of utilizing part of speech tagging[3] is explored for all proposed classification models. In particular, part of speech frequencies of each sentence are included in the input space of the SVM classifiers by including each of the frequencies as an additional input feature/dimension (i.e. 1 new feature for each part of speech, making a total of 36 new features).

---

[2] http://www.lemurproject.org/

## 3. EXPERIMENTAL METHODOLOGY

### 3.1 Dataset

To the best of our knowledge, there is no work done using mathematical word problems so far and therefore we built our own corpora for our experiments. Hence we manually collected 120 mathematical word problems that totally include 518 sentences (out of which 308 are relevant, and 210 are irrelevant) from Grade 3-5 mathematics textbooks [Maletsky et al., 2004] under the guidance and with the help of our collaborators who are experts in educational studies. The problems are grouped with respect to the number of irrelevant sentences within each problem: there are 56 problems having exactly 1, 40 problems having exactly 2, and 24 problems having at least 3 irrelevant sentences. One fourth of the problems in each group are used to train the models and the other three fourths of the problems are used for testing. Details about the relevant/irrelevant sentences are given in Table II.

Table II. Statistics About Each Sentence Type. The number of average terms for each sentence type after stemming and stopwords' removal under the default column; after stemming and avoiding stopwords' removal under the with stopwords column.

| Sentence Type | Number of Total Sentences | Average Length (Words) | |
|---|---|---|---|
| | | default | with stopwords |
| **Relevant** | 308 | 5.42 | 10.39 |
| **Irrelevant** | 210 | 6.03 | 10.34 |

---

[3] http://nlp.stanford.edu/software/tagger.shtml is used.

3.2 Evaluation Metric

To evaluate the effectiveness of the categorization of relevant and irrelevant sentences, we use the common $F_1$ measure, which is the harmonic mean of precision and recall [Baeza-Yates et al., 1999; Rijsbergen, 1979]. Precision (p) is the ratio of the correct categorizations by a model divided by all the categorizations of that model. Recall (r) is the ratio of correct categorizations by a model divided by the total number of correct categorizations. A higher $F_1$ value indicates a high recall as well as a high precision.

$$F_1 = \frac{2pr}{p+r} \qquad (3)$$

4. EXPERIMENT RESULTS

This section presents the experimental results of the methods that are proposed in the Methods and Modeling Approaches section. All the methods were evaluated on the datasets as described in Section 3.1.

An extensive set of experiments was conducted to address the following questions:

1. How effective is the TC method that utilizes the correlation between the question sentence and other sentences along with sentence text (SVM_TermsQSSim) in comparison to the approach that only uses sentence text (SVM_TermsOnly)?
2. What is the effect of stopwords' removal over the performances of all classifiers?
3. What is the effect of utilizing part of speech (POS) tagging over the performances of all classifiers?
4. How effective is the joint probabilistic classification model that additionally utilizes the correlation between all sentences (JointProbClass_Model)?

4.1 The Performance of SVM_TermsQSSim

The first set of experiments was conducted to measure the effect of utilizing the correlation between the questions sentence and other sentences in the classification model. The details about this approach can be found in Section 2.2.

More specifically, SVM_TermsQSSim classifier is compared with SVM_TermsOnly classifier on the categorization of relevant and irrelevant sentences in math word problems. Their performance can be found in Table III and Table IV. It can be seen that the SVM_TermsQSSim classifier almost always outperforms (for 5 out of 6 comparisons; i.e., for 4 comparisons in Table III and 2 additional comparisons in Table IV) the SVM_TermsOnly classifier. The only exception occurs due to the fact that SVM_TermsQSSim classifier loses it advantage of utilizing the correlation between the question sentence and other sentences when stopwords are removed during the correlation estimation in which case estimated correlations are not accurate enough.

Table III. Results of the SVM_TermsQSSim and SVM_TermsOnly classifiers in comparison to each other for two stopword configurations: when stopwords are i) removed or not for the bag of words representation of the input space for the SVM classifiers and ii) when stopwords are removed or not while correlations between the question sentence and other sentences are calculated for each problem. The performance is evaluated with the $F_1$ measure.

| Methods | Stopword Configuration for Terms | | |
|---|---|---|---|
| | default | with stopwords | |
| **SVM_TermsOnly** | 0.592 | 0.668 | |
| **SVM_TermsQSSim** | 0.651 | 0.660 | default |
| | 0.663 | 0.675 | with stopwords |
| | **Stopword Configuration for Similarity Estimation** | | |

Table IV. Results of the SVM_TermsQSSim and SVM_TermsOnly classifiers in comparison to each other for two main configurations for the bag of words representation of the input space for the SVM classifiers: i) when stopwords are removed or not and ii) when part of speech tagging is utilized or not. The performance is evaluated with the $F_1$ measure.

| Methods | Stopword Configuration for Terms | | |
|---|---|---|---|
| | default | with stopwords | |
| **SVM_TermsOnly** | 0.592 | 0.668 | default (without POS) |
| | 0.644 | 0.644 | with POS |
| **SVM_TermsQSSim** | 0.663 | 0.675 | default (without POS) |
| | 0.673 | 0.689 | with POS |
| | **Part-of-Speech Configuration** | | |

Paired t-tests have been applied for this set of experiments and statistical significance with p-value of less than 0.001 has been achieved in 4 out of 6 comparisons in favor of using the correlation between the question sentence and other sentences along with terms over the standard text categorization approach of only using the sentence text. To the best of our knowledge, this is the first work to categorize relevant and irrelevant sentences in math word problems. Results discussed above show that utilizing the correlation between the question sentence and other sentences significantly improves the categorization of relevant and irrelevant sentences in math word problems and the correlation should not be neglected.

4.2 The Effect of Stopwords' Removal

The second set of experiments was conducted to evaluate the effectiveness of including stopwords along with other words i) as terms that constitute the feature space of the SVM models, and ii) as terms that constitute the feature space to be used by the Cosine similarity to measure the similarity between sentences (or term vectors) in a word problem. The details about this approach are given in Section 2.4.

This section specifically compares all classifiers with themselves with respect to their performances when stopwords are included or not as additional dimensions to their feature spaces. It can be seen from the results in Table III that SVM_TermsOnly classifier has a statistically significant performance improvement (with p-value less than 0.001) when stopwords are included in the feature space of the SVM_TermsOnly classifier. For the SVM_TermsQSSim classifier, inclusion of stopwords are two-fold, as discussed above, i) they can be incorporated into the feature space of their SVM models (note that 2 comparisons in Table III and 1 more comparison in Table IV can be made), ii) they can be incorporated into the feature space of Cosine similarity (note that 2 comparisons can be made in Table III). For the second configuration (i.e., when stopwords are incorporated into the feature space of Cosine similarity or not), statistically significant results have been achieved (with p-value less than 0.01) in favor of including the stopwords into the feature space of Cosine similarity for both comparisons. For the first configuration, statistically significant results (with p-value less than 0.05) have again been achieved in favor of including the stopwords into the feature space of SVM models for 2 out of 3 comparisons and these results are consistent with the prior work on the categorization of mathematical word problems with respect to their types [Cetintas et al., 2009]. The only exception occurs due to the fact that SVM_TermsQSSim classifier behaves relatively bad when stopwords are removed during the correlation estimation in which case estimated correlations are not accurate enough (note the same exception discussed in Section 4.1). One interesting observation for this set of experiments is that completely ignoring stopwords in either case does not lead to a sharp reduction on the performance of SVM_TermsQSSim classifier due to the fact that incorporation of the sentence to question sentence similarities into the SVM model becomes effective enough for resulting in a superior performance over the configuration of the SVM_TermsOnly classifier that eliminates stopwords.

## 4.3 The Performance of Utilizing Part of Speech (POS) Tagging

The third set of experiments was conducted to evaluate the effectiveness of utilizing part of speech tagging (i.e., including part of speech frequencies) into the feature space of the feature space of the SVM models. The details about this approach are given in Section 2.5.

This section specifically compares all classifiers with themselves with respect to their performances when part of speech frequencies are included or not as additional dimensions to their feature spaces. It can be seen from the results in Table IV and V that utilization of part of speech tagging does not always improve the categorization performance. Statistical significance (with p-value less than 0.001) has been achieved for some of the comparisons both in favor of and against using part of speech tagging. Specifically, statistical significance (with p-value less than 0.05) has been achieved in favor of using POS tagging for SVM_TermsQSSim classifier for both stopword configurations in Table IV. Similarly, significant improvement has been observed (with p-value much less than 0.001) for SVM_TermsOnly classifier for the "default" configuration of stopwords in Table IV whereas significantly worse (with p-value less than 0.001) results have been observed for the "with stopwords" configuration in Table IV. For the JointProbClass_Model classifier, the observed results that can be seen in Table V have been found to be not significantly different (i.e., p-value bigger than 0.05) in favor of or against using POS tagging. Overall, utilizing POS tagging has not been found to be an effective approach to improve the effectiveness of the categorization although it has been shown to be effective for the categorization of math word problems with respect to their types in [Cetintas et al., 2009]. This can be explained by the fact that almost all of the POS frequencies (except for NNPN & WRP) are uniformly distributed among the relevant and irrelevant sentences in math word problems as it can be seen in Figure 2. Indeed, WRP tag covers some discriminative stopwords like "how" that can mostly be seen in relevant sentences. This helps the SVM_TermsOnly classifier alleviate the performance loss in case of stopwords removal (i.e., the "default" stopword configuration in Table IV). Yet, when stopwords are included, POS tagging again becomes ineffective.

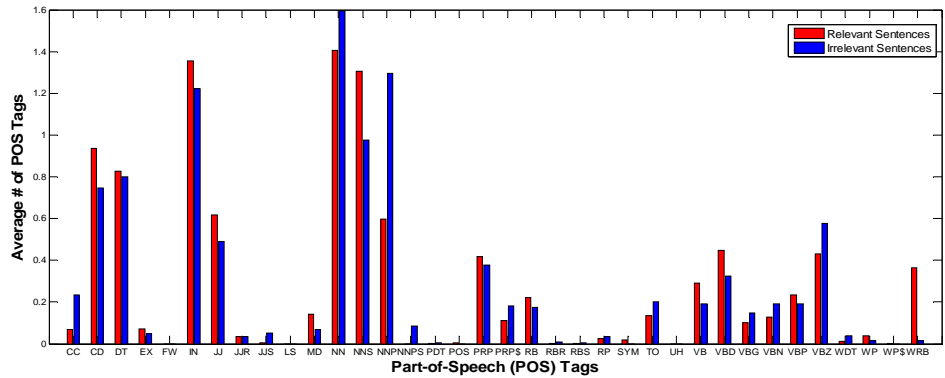

Fig. 2. Distribution of parts of speech across relevant and irrelevant sentences (frequencies averaged per sentence) in math word problems.

## 4.4 The Performance of the Joint Probabilistic Classification Model (JointProbClass_Model)

The final set of experiments was conducted to evaluate the effectiveness of incorporating the correlations between all sentences into the classification models, namely the joint probabilistic classification model. The details about this approach are given in Section 2.3.

Table V. Results of the JointProbClass_Model, SVM_TermsQSSim and SVM_TermsOnly classifiers in comparison to each other for two configurations: when part of speech tagging is i) utilized and ii) not utilized for the input space of the SVM classifiers. The performance is evaluated with the $F_1$ measure.

| Methods | Part-of-Speech Configuration | |
|---|---|---|
| | with POS | without POS |
| **SVM_TermsOnly** | 0.644 | 0.668 |
| **SVM_TermsQSSim** | 0.689 | 0.675 |
| **JointProbClass_Model** | 0.710 | 0.716 |

Particularly JointProbClass_Model classifier is compared with SVM_TermsQSSim classifier (with the configuration that includes stopwords in the feature space used during the correlation estimation) and SVM_TermsOnly classifier. The performance of JointProbClass_Model classifier is shown in Table V. It can be seen that the JointProbClass_Model classifier outperforms both the SVM_TermsOnly classifier and the SVM_TermsQSSim classifier for both POS configurations. Statistical significance (with p-value much less than 0.001) has been achieved in favor of utilizing the correlation between all sentences for all four comparisons in Table V. It should be noted that the improvement gained by using the correlation between all sentences help to achieve the highest and most significant improvement among all models, and configurations (except the performance jump by the approach of including stopwords into the feature space of the SVM_TermsOnly classifier over the weak baseline). This set of experiments explicitly shows that although utilizing the correlation between the question sentence and other sentences in a problem is highly beneficial to better see the relationship between relevant sentences in the problem, it is not enough when some of the relevant sentences do not have too much in common with the question sentence. Therefore, a joint classification model in which the correlations between all sentences are taken into account becomes significantly much more effective. Regarding the fact that irrelevant sentences do not have much in common with the question sentence, this joint model is able to capture the relationship between irrelevant sentences as well as the relationship between relevant sentences. Prior work on categorization of relevant and irrelevant questions asked in a micro-blogging supported classroom showed the effectiveness of utilizing the correlation between the questions and available lecture materials [Cetintas et al., 2010]. Since i) there is no external source that can be utilized for the categorization of sentences in math word problems, and ii) there are much less sentences in a math word problem than questions asked in a lecture, a joint probabilistic model becomes tractable to capture the correlations between all sentences in math word problems. Experiments results demonstrate the effectiveness of utilizing a joint probabilistic classification approach as well as utilizing the correlations between all sentences.

## 5. CONCLUSIONS, DISCUSSION AND FUTURE WORK

Categorization of relevant and irrelevant sentences in math word problems is an important step for estimating the difficulty level of math word problems [Marzocchi et al., 2002; Mastropieri and Scruggs, 2006; Passolunghi and Siegel, 2001] and this is an important task for many applications. Systems that use problem posing for accelerating content authoring or for deepening students' understanding need to automatically assess the difficulty level of problems posed by students (especially for being able to model students' problem posing performances) [Birch and Beal, 2008]. Systems that try to automate the process of problem solving (e.g., STUDENT system) need to deal with verbose problems and irrelevant information [Bobrow, 1964]. Similarly, systems that try to automate the process of building problem libraries of intelligent tutoring systems by analyzing the math word problems (e.g., by categorizing their types as in [Cetintas et al., 2009]) need to also analyze the difficulty levels of the problems before the problems become usable.

This paper proposes a novel application of text categorization to identify relevant and irrelevant sentences in math word problems. Several modeling approaches and several preprocessing configurations are studied for this application through extensive experiments. Empirical results show that utilizing the correlation between the question sentence and other sentences is significantly more effective than using sentence text only. Similarly, utilizing the correlations among all sentences (along with the correlation between the question sentence and other sentences and sentence text) has been found to be significantly more effective than i) the approach of utilizing the correlation between the question sentence and sentence text and ii) the approach that only uses sentence text. Furthermore, it is found to be significantly more effective to include stopwords into the feature space of SVM classifiers as well as into the feature space to be used by the Cosine measure during the correlation estimation. Finally, utilizing part of speech tagging is found not to be effective, although it has been shown to be useful for categorizing mathematical word problems with respect to their types in prior work.

There are several possibilities to extend the research. The first direction for potential future work results from the fact that the accuracy of the proposed classifiers is not high enough to utilize directly. It should be noted that this is the first work on categorization of relevant and irrelevant sentences in mathematical word problems, and several state-of-the-art text categorization techniques have been used as well as several novel approaches that have been proven to be effective. The fact that the accuracy of the best classifier is still not high enough is mainly due to three reasons. Firstly, sentences are very short (as shown in Table II), and this makes the classification task substantially harder due to data sparsity. Secondly, although both relevant and irrelevant sentences are correlated among themselves more, they are also correlated with each other since all of them give information about the same problem within the same context. This correlation makes the classification task significantly harder as well. These challenges have also been observed in the recent related study on identifying relevant and irrelevant questions in a micro-blogging supported classroom [Cetintas et al., 2010; Cetintas et al., To Appear]. Thirdly, only a small amount of training data has been used in this study. Therefore, it is worthwhile to see the effect of using more training data on the performance of the classifiers. On the other hand, it is possible to use the decisions of the current classifiers with confidence scores (i.e., the decisions that the classifiers are confident enough can be used directly whereas the decisions that the classifiers are not confident enough can be forwarded for human judgment). The second direction for potential future work is due to the fact that the SVM classifiers in this work are using single words as their features (i.e., unigram model). It may be helpful to explore n-gram models. However, it should be noted that bi-gram or n-gram models are harder to learn (due to much larger feature space). With the fact that the sentences are short and the number of available math problems is limited, overfitting will be a problem that will have to be dealt with. Yet, it is still worthwhile to explore this direction in detail in a separate throughout study.

## ACKNOWLEDGEMENTS

This research was partially supported by the NSF grants IIS-0749462, IIS-0746830 and DRL-0822296. Any opinions, findings, conclusions, or recommendations expressed in this paper are the authors', and do not necessarily reflect those of the sponsor.

## REFERENCES

ARROYO, I. & WOOLF, B. P. 2003. Students in AWE: changing their role from consumers to producers of ITS content. In *Proceedings of the 11th International AIED Conference, Workshop on Advanced Technologies for Math Education*.

ARROYO, I., SCHAPIRA, A. & WOOLF, B. P. 2001. Authoring and sharing word problems with AWE. In *Proceedings of the 10th International AIED Conference*.

BAEZA-YATES, R. AND RIBEIRO-NETO, B. 1999. *Modern information retrieval*. ACM Press Series/Addison Wesley, 75-82.

BEAL, C. R. 2007. On-line tutoring for math achievement testing: A Controlled Evaluation. In *Journal of Interactive Online Learning,* 6(1):43-55.

BIRCH, M., & BEAL, C. R. 2008. Problem posing in AnimalWatch: an interactive system for student-authored content. In *Proceedings of the 21st International FLAIRS Conference*.

BOBROW, D. 1964. Natural Language Input for a Computer Problem Solving System. PhD Thesis, MIT.

BROWN, S. I. & WALTER, M. I. 1990. *The art of problem posing*. Hillsdale NJ: Lawrence Erlbaum.

CETINTAS, S., SI, L., CHAKRAVARTY, S., AAGARD, H. P., BOWEN, K. 2010. Learning to identify students' relevant and irrelevant questions in a micro-blogging supported classroom. In *Proceedings of ITS-10, the 10th International Conference on Intelligent Tutoring Systems,* 281-284.

CETINTAS, S., SI, L., AAGARD, H. P., BOWEN, K., & CORDOVA-SANCHEZ, M. (To Appear). Micro-blogging in classroom: Classifying students' relevant and irrelevant questions in a micro-blogging supported classroom. In *IEEE Transactions on Learning Technologies (Accepted October 2010).*

CETINTAS, S., SI, L., XIN, Y. P., ZHANG, D., PARK, J. Y. 2009. Automatic text categorization of mathematical word problems. In *Proceedings of the 22nd International FLAIRS Conference,* 27-32.

COWELL, R. G., DAWID, A. P., LAURITZEN, S. L. & SPIEGELHALTER, D. J. 1999. In *Probabilistic networks and expert systems.* Springer.

FRAKES, W. and BAEZA-YATES, R. 1992. *Information retrieval: data structures and algorithms.* Prentice Hall, Englewood Cliffs, NJ.

GOLLUP, J.P., BERTENTHAL, M., LABOV, J., CURTIS, P. C. 2002. *Understanding: improving advanced study of mathematics and science in US high schools.* National Academy Press, Washington, DC.

HINTON, G. and SEJONWSKI, T. 1986. *Learning and relearning in Boltzmann machines.* In Rumelhart, editor, Parallel Distributed Processing, 282–317. MIT Press.

HIRASHIMA, T., YOKOYAMA, T, OKAMOTO, M., & TAKEUCHI, A. 2007. Learning by problem-posing as sentence-integration and experimental use. In *Proceedings of the 13th International AIED Conference*.

JOACHIMS, T. 1998. Text categorization with support vector machines: learning with many relevant features. In *Proceedings of ECML-98, 10th European Conference on Machine Learning*.

JOACHIMS, T. 1999. *Making large-scale support vector machine learning practical.* MIT Press.

KO, J. and NYBERG, E. and SI, L. 2007. A probabilistic graphical model for joint answer ranking in question answering. In *Proceedings of ACM SIGIR-07, 30th ACM SIGIR Conference on Research and Development in Information Retrieval,* 343-350.

KOEDINGER, K. R., ANDERSON, J.R., HADLEY, W.H., & MARK, M. A. 1997. Intelligent tutoring goes to school in the big city. In *International Journal of Artificial Intelligence in Education,* 8, 30-43.

MALETSKY, E. M., ANDREWS, A. G., BURTON, G. M., JOHNSON, H. C., LUCKIE, L. A. 2004. *Harcourt Math (Indiana Edition).* Chicago: Harcourt.

MARZOCCHI, G. M., LUCANGELI, D., MEO, T. D., FINI, F., CORNOLDI, C. 2002. The disturbing effect of irrelevant information on arithmetic problem solving in inattentive children. In *Developmental Psychology*, 21(1): 73-92.

MASTROPIERI, M. A. & SCRUGGS, T. E. 2006. *The inclusive classroom: Strategies for effective instruction.* New Jersey: Prentice Hall.

MINKA, T. 2003. *A comparison of numerical optimizers for logistic regression.* Unpublished draft.

PASSOLUNGHI, M. C. AND SIEGEL, L. S. 2001. Short-term memory, working memory, and inhibitory control in children with difficulties in arithmetic problem solving. In *Journal of Experimental Child Psychology,* 80(1): 47-57.

PORTER, M. F. 1980. An algorithm for suffix stripping. In *Program: Electronic Library and Information Systems*, 14(3):130–137.

RIJSBERGEN, C. J. v. 1979. *Information retrieval, 2nd edition.* University of Glasgow.

RITTER, S., ANDERSON, J., CYTRYNOWICZ, M., & MEDVEDEVA, O.1998. Authoring content in the PAT algebra tutor. In *Journal of Interactive Media in Education*, 98(9).

SCOTT, S. and MATWIN, S. 1999. Feature engineering for text classification. In *Proceedings of ICML-99, 16th International Conference on Machine Learning*.

SEBASTIANI, F. 2002. Machine learning in automated text categorization. In *ACM Computing Surveys*, 34(1):1–47.

SHIAH, R., MASTROPIERI, M. A., SCRUGGS, T.E., & FULK, B..M. 1995. The effects of computer-assisted instruction on the mathematical problem solving of students with learning disabilities. In *Exceptionality*.

U.S. DEPT. OF EDUCATION. 2006. No Child Left Behind is working. *Retrieved November, 2008, from* http://www.ed.gov/nclb/overview/importance/nclbworking.html

VERSCHAFFEL, L., GREER, B., CORTE, D. E. 2000. *Making sense of word problems.* Taylor & Francis.
YANG, Y. & LIU, X. 1999. A re-examination of text categorization methods. In *Proceedings of SIGIR-99, 22nd ACM SIGIR Conference on Research and Development in Information Retrieval*, 42–49.